\documentclass[10pt, conference, compsocconf]{IEEEtran}
\ifCLASSINFOpdf
   \usepackage[pdftex]{graphicx}
\else
\fi
\usepackage[cmex10]{amsmath}
\usepackage[caption=false,font=footnotesize]{subfig}
\usepackage{stfloats}
\usepackage{xcolor}
\usepackage{amsmath}
\usepackage{amsfonts} % BSM
\usepackage[super]{nth}
\usepackage{float}
\usepackage{comment}
\usepackage{multirow}
\usepackage{pifont}
\usepackage[bookmarks=false]{hyperref}
\newcommand{\stimes}{{\times}}

\begin{document}
%
% paper title
% can use linebreaks \\ within to get better formatting as desired
\title{Deep Visual Template-Free Form Parsing}

% author names and affiliations
% use a multiple column layout for up to two different
% affiliations

\author{
\IEEEauthorblockN{Brian Davis, Bryan Morse}
\IEEEauthorblockA{Brigham Young University\\
\{briandavis, morse\}@byu.edu}
\and
\IEEEauthorblockN{Scott Cohen, Brian Price, Chris Tensmeyer}
\IEEEauthorblockA{Adobe Research\\
\{scohen,bprice,tensmeye\}@adobe.com}
}

% conference papers do not typically use \thanks and this command
% is locked out in conference mode. If really needed, such as for
% the acknowledgment of grants, issue a \IEEEoverridecommandlockouts
% after \documentclass

% for over three affiliations, or if they all won't fit within the width
% of the page, use this alternative format:
% 
%\author{\IEEEauthorblockN{Michael Shell\IEEEauthorrefmark{1},
%Homer Simpson\IEEEauthorrefmark{2},
%James Kirk\IEEEauthorrefmark{3}, 
%Montgomery Scott\IEEEauthorrefmark{3} and
%Eldon Tyrell\IEEEauthorrefmark{4}}
%\IEEEauthorblockA{\IEEEauthorrefmark{1}School of Electrical and Computer Engineering\\
%Georgia Institute of Technology,
%Atlanta, Georgia 30332--0250\\ Email: see http://www.michaelshell.org/contact.html}
%\IEEEauthorblockA{\IEEEauthorrefmark{2}Twentieth Century Fox, Springfield, USA\\
%Email: homer@thesimpsons.com}
%\IEEEauthorblockA{\IEEEauthorrefmark{3}Starfleet Academy, San Francisco, California 96678-2391\\
%Telephone: (800) 555--1212, Fax: (888) 555--1212}
%\IEEEauthorblockA{\IEEEauthorrefmark{4}Tyrell Inc., 123 Replicant Street, Los Angeles, California 90210--4321}}

% use for special paper notices
%\IEEEspecialpapernotice{(Invited Paper)}

% make the title area
\maketitle

\begin{abstract}
Automatic, template-free extraction of information from form images is challenging due to the variety of form layouts. 
This is even more challenging for historical forms due to noise and degradation.
A crucial part of the extraction process is associating input text with pre-printed labels. 
We present a learned, template-free solution to detecting pre-printed text and input text/handwriting and predicting pair-wise relationships between them.
While previous approaches to this problem have been focused on clean images and clear layouts, we show our approach is effective in the domain of noisy, degraded, and varied form images. 
We introduce a new dataset of historical form images (late 1800s, early 1900s) for training and validating our approach.
Our method uses a convolutional network to detect pre-printed text and input text lines.
We pool features from the detection network to classify possible relationships in a language-agnostic way. 
We show that our proposed pairing method outperforms heuristic rules and that visual features are critical to obtaining high accuracy.
\end{abstract}


% For peer review papers, you can put extra information on the cover
% page as needed:
% \ifCLASSOPTIONpeerreview
% \begin{center} \bfseries EDICS Category: 3-BBND \end{center}
% \fi
%
% For peerreview papers, this IEEEtran command inserts a page break and
% creates the second title. It will be ignored for other modes.
\IEEEpeerreviewmaketitle

%%%%%%%%%%%%%%%%%%%%%%%%%%%%%%%%%%%%%%%%%%%%%%%%%%%%%%%%%%%%%%%%%%%%%%%%%%%%

\section{Introduction}

Forms are a long-used and convenient device for collecting information. % from people. 
However, in modern times we prefer to have data stored in digital databases rather than physical archives.
Extracting the information from images of forms into databases is a problem confronting both businesses and those interested in preserving history.

\begin{figure}[!t]
\centering
\includegraphics[width=0.48\textwidth]{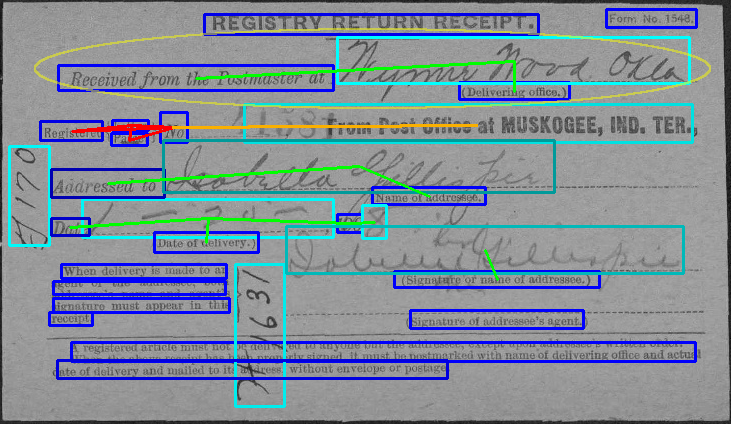}
\caption{Example label-value relationship pairing. We detect pre-printed text (blue boxes), input text (cyan boxes), and label-value relationships (colored lines) in historical form images. 
Line colors indicate the correctness of the pairing: true-positive (green), false-negative (orange), or false-positive (red). 
Note that some relationships are not one-to-one, such as the instance inside the yellow ellipse.
%\vspace{-0.125in}
}
\label{fig:top_fig}
\end{figure}

This work focuses on the problem of detecting pre-printed text and input text (handwritten/stamped/typed text added to the form) in a noisy form image and determining which text instances should be paired, as shown in Fig.~\ref{fig:top_fig}.
When extracting information from a form, knowing the semantic meaning of the input text is often as important as knowing its transcription.
Typically, label-value relationships exist between certain pre-printed text and input text elements in a form, and the input text's semantic meaning can be inferred from the label. 
%We indirectly find comments as those handwriting/fill instances not paired with any text. 
In some instances these relationships are not exclusively one-to-one,
as illustrated at the top of Fig.~\ref{fig:top_fig}.
%for example, the input text at the top of Fig.~\ref{fig:top_fig} (yellow ellipse) has two pre-printed text instances related to it that collectively indicate its semantic meaning. 
%Additionally, in future work when relationships between pre-printed text and other pre-printed text are considered (e.g., header-content relationships), there will be many additional one-to-many relationships.

%While reading the text and being able to discern relationships between them at a semantic level is a helpful step in this pairing problem, 

Our method is language-agnostic and does not use text transcriptions, meaning our method can directly be applied to forms in different languages if visual characteristics are the same. 
%While having transcriptions may make relationships easier to determine, forms are typically designed to be clear in their label-value associations from a visual perspective.
While transcriptions may make relationships easier to determine, label-value relationships in forms are typically clear from a purely visual perspective.
For example, most people can view a form in an unfamiliar language and infer the label-value relationships. %which input text is associated with which pre-printed text.
%We choose to ignore ATR in this work. 
%for two primary reasons. First, ignoring the content of text makes the method language agnostic. Second, it can be observed that forms are typically designed to be clear in their label-value associations from a visual perspective, that is, most people can view a form in an unfamiliar language and infer which text is associated with which fields.
%(really the primary reason is we don't have the transcriptions...)
% BSM: commenting out this note since we're not emphasizing the global optimization
%\textcolor{red}{ add example if room (I made one for FHTW using lorem ipsum) which also shows the need of global optimization}
We propose that a deep neural network should be able to infer these relationships as well.

We detect text lines using a Fully Convolutional Network (FCN) as it is effective, simple, and provides features for later processes to reuse.
%We restrict relationships considered using a line-of-sight heuristic rather than considering all possible relationships. 
We use a convolutional classifier network to predict which potential relationships are correct using a context window around the relationship.
To ensure globally coherent solutions, we predict the number of neighbors for each text line and apply an optimization procedure to find the best set of relationships based on relationship probabilities and the predicted number of neighbors.

Our primary contribution is a trained, end-to-end, language-agnostic method for finding label-value pairs in noisy, novel form images that outperforms heuristic pairing methods. % as well as learned methods trained on non-visual features.
We also show that using dilated, non-square kernels in the FCN text detector improves detection accuracy for long text lines.
%This method includes a FCN architecture for detecting long text lines.
Finally, we contribute a new annotated dataset of historical form images, the National Archives Forms (NAF) dataset.
Our code is at \url{http://github.com/herobd/Visual-Template-Free-Form-Parsing} and the NAF dataset is available at \url{http://github.com/herobd/NAF_dataset}.

\begin{comment}
We also present two other contributions:
\begin{itemize}
  \item A fully convolutional architecture for detecting long text lines.
  %\item A method for optimizing pairing results based on a prior of the number of neighbors each label or value has.
  \item A new annotated dataset of historical form images, the National Archives Forms dataset.
\end{itemize}
\end{comment}

%%%%%%%%%%%%%%%%%%%%%%%%%%%%%%%%%%%%%%%%%%%%%%%%%%%%%%%%%%%%%%%%%%%%%%%%%%%%
\section{Previous Work}

%The problem of extracting information from form images has a rich history of research and almost all methods address the following four problems \cite{barrett2004digital,butt2012information,zhou2016irmp,hammami2015one,hirayama2011development}:
%\begin{enumerate}
% 
%    \item Detecting pre-printed text and input text.
%    \item Automated Text Recognition (ATR)\footnote{Optical Character Recognition and/or Handwritten Text Recognition}of input text %and possibly pre-printed text.
%    \item Identifying the semantic meaning of the input text.
%    \item Formatting appropriately for its semantic meaning the recognized content of the input text.
%\end{enumerate}
%Each of these has their own sub-problems and can be closely tied with each other. 
%Additionally, inferring the data schema may be an important step in template-free extraction.

%Our work focuses on detection of text elements (Problem~1) and the visual linking of pre-printed and input text (part of Problem~3).
While a complete solution to the problem of extracting information from forms has many parts (text detection, text recognition, determining the semantic meaning of text, etc.), we review here the aspects most related to our focus, namely detecting and pairing pre-printed and input text.

\subsection{Text Detection}
%In template-based form processing, detecting the text is generally equivalent to aligning a template to the image~\cite{barrett2004digital}. 
%In template-free processing we must detect text without prior knowledge of the form's layout.

Older methods for text detection have largely focused on free-form documents, rather than forms and other documents with complex layouts.
They have used projection profiles, smearing, or bottom-up methods to identify text lines, all of which must be aware of text regions a priori. They are also not generally resistant to noise.
For a survey of these and similar techniques, we refer the interested reader to~\cite{likforman2007text}.  % How's this?
%\textcolor{red}{All of my information from this paragraph is from the survey referenced at the end. Do I need to cite specific papers for the other methods?  Bryan: Yes.}
%\nocite{gruning2018}

Modern approaches have overcome these obstacles using deep learning, presenting solutions that are robust in the presence of noise, arbitrary document layouts, arbitrary orientation, and curved text lines. %~\cite{gruning2018,wigington2018start}. 
Gr{\"u}ning et al.~\cite{gruning2018} use a FCN for pixel labeling followed by post-processing to extract text lines from the pixel predictions.
Wigington et al.~\cite{wigington2018start} use a FCN to detect the beginning of text lines and have a network segment the line by stepping along it. 
%They have line termination tied to ATR, although it could be extended to predict the termination points explicitly.
% join the paragraphs?
%Either of these methods would be suitable for our problem and 
Like these methods, we use a FCN, but our method is simpler as it directly predicts bounding rectangles.
This limits the types of text lines we can detect (straight, horizontal), but is suitable for our dataset.

\subsection{Form Processing}

Much of the previous work in form processing assumes the availability of templates for form types of interest~\cite{barrett2004digital,butt2012information,hammami2015one}. %This template is aligned to the image and the fields, with known meaning from the template, can be accurately found even in the presence of some noise \cite{barrett2004digital}. 
%Later work \cite{butt2012information} makes the inclusion of such templates easier by automatically learning the layout from an annotated blank form image.
%Though having a template is practical in some contexts, in other contexts it is too restrictive and burdensome. One example of this would be the automatic processing of a historical archive where the documents are not sorted and assumptions about what form types might be present can't be made. Another would be a business providing form processing as a service, where they are not tied to the process of new forms being created.
%This assumption has been relaxed in later work where a table format, cells, or other regularity can be assumed~\cite{zhou2016irmp,hirayama2011development}.
%\textcolor{red}{TODO: List how each does pairing. No common dataset. Actually list out each assumptions. Provide example images to contrast with our data.}
%zhou2016irmp uses cells
This assumption has been relaxed in later work \cite{zhou2016irmp,hirayama2011development}.
Zhou et al.~\cite{zhou2016irmp} assume all relevant information is contained in table structures with lines that can be detected by an OCR engine. Many forms, however, do not have table structures.
The method of Hirayama et al.~\cite{hirayama2011development} is more general as it allows greater variation in layout. 
It scores potential labels (pre-printed text) and values (input text) by matching transcribed text with predefined class-dependent dictionaries and rules. % according to rules for certain classes. 
Possible relations between text instances are scored using heuristic layout rules.
The combination of these two scores %, likely layout orientation and matching classes between text instances, 
define their final pairing score. 
Hirayama et al.~\cite{hirayama2011development} focus on extracting the subset of information from the forms described by the predefined text dictionaries/rules.

Many assumptions made by these methods are broken in our proposed NAF dataset of historical forms.
%Several of the assumptions these methods make are broken in the dataset of historical forms we evaluate our results on.
The high noise levels in historical forms can affect the accuracy of classical layout analysis (e.g., line detection).
%For historical forms, noise levels can be very high which can effect layout anaylsis such as detecting lines. 
While~\cite{hirayama2011development} handles varied layouts, we show that their heuristic layout rules do not generalize well to the NAF dataset.

We are unaware of any publicly available datasets or official reference implementations of prior work that would enable direct comparison with our proposed method.
%We note that no previous methods we are aware of evaluate on publicly available datasets.
%(e.g., having text flow outside of a box).
%Previous methods to solve the problem on converting filled in form images to digital information have relied on assumptions of either the existence of templates or reasonably clear guides.

%TODO figure contrasting domains

% BSM: artificially placing this here to cause it to appear after the next page break and in the appropriate section
\begin{figure*}
\centering
\includegraphics[width=0.99\textwidth]{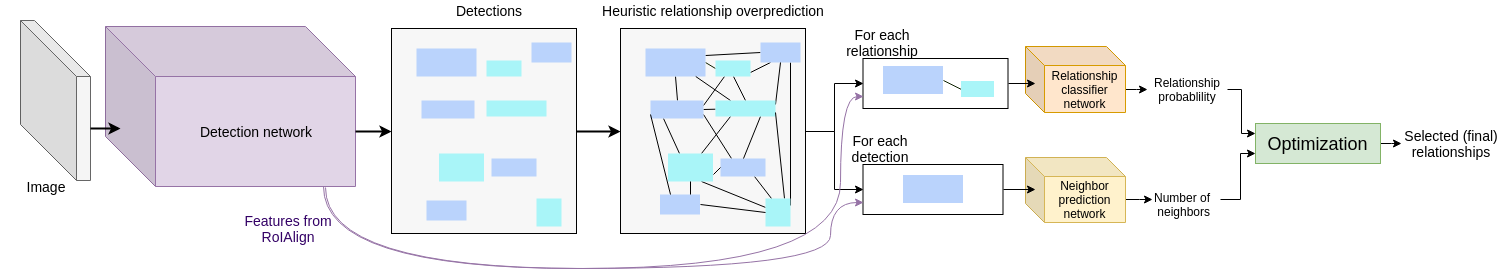}
\caption{Overview of our method. First we detect pre-printed text and input text lines (shown in more detail in Fig.~\ref{fig:detectornet}). We then find possible relationships using a line-of-sight heuristic.
We then take features from context windows around each detection and each relationship with RoIAlign. These are fed through separate convolutional networks to predict the number of neighbors and the probability of the relationship being true.
These predictions are then passed to a global optimization to produce the final relationship predictions.}
\label{fig:overview}
\end{figure*}

\subsection{Scene Graphs}

The problem of finding label-value pairs in a form is closely related to the problem of creating scene graphs from natural images, and our proposed method is similar to some previous work in this domain.
In scene graphs, the objects in the image are nodes, and edges represent relationships (``on top of'', ``is part of'', etc.) between the objects. 
For our work on forms, pre-printed and input text instances are the objects, and we only consider the label-value relationship.
%In forms the objects are the text instances; in this work we are concerned only with label-value relationships. 
%Our method is similar to previous work in this domain.

%Almost all methods begin by detecting and classifying the objects in an image, hypothesizing relationships, and then refining and classifying those relationships.

Zhang et al.~\cite{zhang2017relationship} use a detection network to predict object and relationship bounding boxes. 
% over the image detecting objects and relationships (the rectangle including both objects in the relationship). 
%They prune some relationships based on whether the locations are spatially coherent.
%Then they potential relationships based on learned visual features of object pairs and spatial features computed from the detected bounding box geometries.
Then they combine two scores (late fusion) for determining which relationships should be kept.
One score is based on learned visual features of object pairs, and the other score is based on spatial features computed from the detected bounding box geometries.
%One score is produced using region of interest pooling over the three detections (object, subject, and relationship) and combining the three pooled areas so a small network can compute the score. 
%The other score is based on the differences in location and shape of the three detections.
%and classifies based on this.
We take a similar approach, but perform early fusion of visual and spatial features by inputting them to our network, and we use a heuristic to generate candidate relationships instead of a learned network.

Yang et al.~\cite{yang2018graph} initially only detect and classify objects, but later use a relation-proposal network to predict relationships based on object classes. %that uses the predicted classes of the objects to hypothesize possible relationships. 
The proposed relationships are formed into a graph and an attention graph convolution network predicts final relationship and object classes.
%It then creates a graph using these proposed relationships and uses an attention graph convolution network to produce the edge classifications and final object classifications.
LinkNet~\cite{woo2018linknet} also uses a relation-proposal network and produces object embeddings that are used to find compatible objects for each type of relationship.
%This is used to further refine both the object-level predictions as well as the relationship predictions.
%\textcolor{red}{I didn't follow this paper when I read it earlier.}
% BSM: read Woo paper and took a shot at summarizing it.

%Our method is similar to \cite{zhang2017relationship}, although we do not detect relationships but instead propose them through a heuristic metric. 
%Unlike \cite{zhang2017relationship} we do not have separate visual and spatial scores, but include both visual and spatial features into our relationship classifier.
%Unlike \cite{zhang2017relationship}, our RoI around both objects in a relationship is padded for context and includes masks for the locations of detections.

%%%%%%%%%%%%%%%%%%%%%%%%%%%%%%%%%%%%%%%%%%%%%%%%%%%%%%%%%%%%%%%%%%%%%%%%%%%%

% figure fig:overview really belongs here

\section{Method Overview}
\begin{figure*}
    \centering
    %\begin{subfigure}[]{}
    \subfloat[Using normal $3 \stimes 3$ convolutions]{
        \includegraphics[width=0.49\textwidth]{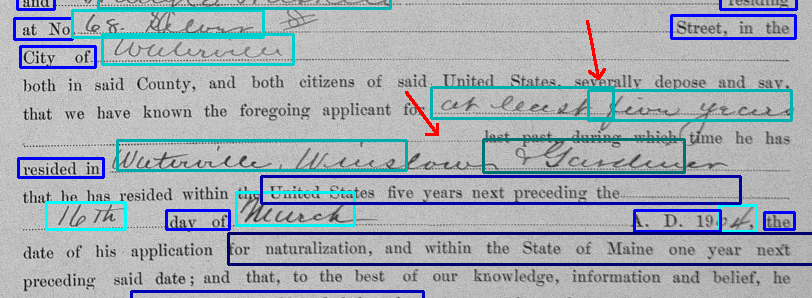}
    }
    \subfloat[Using dilated $1 \stimes 3$ convolutions]{
        \includegraphics[width=0.49\textwidth]{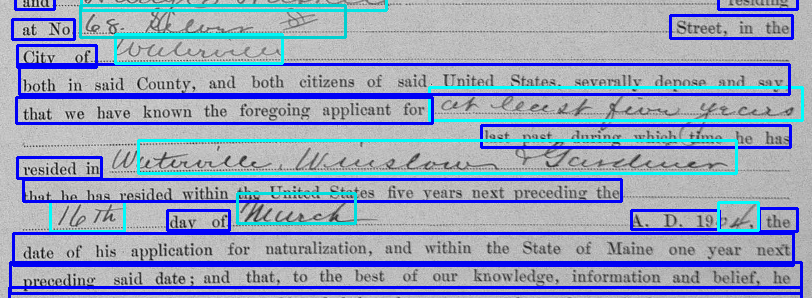}
    }
        %\caption{A tiger}
        %\label{fig:tiger}
    %\end{subfigure}
    \caption{Detection without dilation (a) and using dilated $1 \stimes 3$ convolutions (b). Blue boxes are pre-printed text detections, cyan boxes are input text detections. Notice in (a) that long lines are either broken into multiple detections (red arrows) or missed.
    %\vspace{-0.14in}
    }
    \label{fig:ex_detect}
\end{figure*}

An overview of our method to find label-value pairs in form images can be seen in Fig.~\ref{fig:overview}. 
We detect text/handwriting instances and find possible relationships using a line-of-sight heuristic. 
For each possible relationship, we extract features from the detection network around the two text instances (padded for context).
These features and detection location masks are fed to a small convolutional network to predict the probability of the relationship.

We also perform a global optimization, which takes the predicted probabilities of the relationships and a predicted number of neighbors (number of relationships) per detection and selects which relationships are most in agreement with both of these sets of predictions.

To predict each detection's number of neighbors, the detector network first predicts an initial estimate. 
To refine this estimate, we mimic our process of predicting relationships, focusing on a single detection rather than a pair.
%by extracting features around each detection (with context) and augment these features with masks of its own and other detections' locations. 
%This input is fed to a small convolutional network that predicts the number of neighbors for the detected element.

%%%%%%%%%%%%%%%%%%%%%%%%%%%%%%%%%%%%%%%%%%%%%%%%%%%%%%%%%%%%%%%%%%%%%%%%%%%%
\section{Detection}
We frame the problem of detecting pre-printed text and input text as object detection and use an FCN approach, similar to YOLOv2 \cite{yolov2}, to predict text line bounding boxes and classes (pre-printed text or input text). 
We show that FCNs with dilated $1 \stimes 3$ convolutions detect long text lines as single entities significantly better than FCNs that use only un-dilated $3 \stimes 3$ convolutions.
We choose to detect text at the line level as this is the input expected by state-of-the-art handwriting recognition methods~\cite{puigcerver2017multidimensional}. 
%An additional step a full extraction method would need to address is identifying which lines belong to the same entity, but we omit this step from our work and focus on single-line text labels for input fields.

For most forms the printed text and handwriting are reasonably horizontal. 
The primary exception is comments, which are often oriented independently of the document. 
While work has been done to detect accurate bounding regions for skewed and even curved text \cite{wigington2018start,gruning2018}, we choose a simpler method that is robust to small amounts of skew and assumes straight lines. 
%removed refs to liao2018rotation,ma2018arbitrary
 
Our approach is based on YOLOv2 \cite{yolov2} and uses the loss formulation of YOLOv3 \cite{yolov3}.
This model uses a FCN and at each position predicts the probability of objects being present for a number of anchor boxes (prior shapes), how the anchors should be changed to align with the object, and the class of the object. % (text or input text in our case).
%Previous works have used fully-convolutional architectures for finding the start of text/handwriting lines \cite{wigington2018start,moysset2017full}, but rely on other methods to extract the whole line.
%For our dataset we find direct bounding box prediction effective, given some specific architecture modifications.

%While Faster-RCNN \cite{rcnn} is often the baseline detection model chosen in other works \cite{zhang2017relationship,yang2018graph,woo2018linknet}, we choose the YOLO model as it is lighter, particularly in lacking a separate classifying network. As we have only two classes, we feel it is unnecessary to need a separate network to classify.

\begin{figure}
\centering
\includegraphics[width=0.49\textwidth]{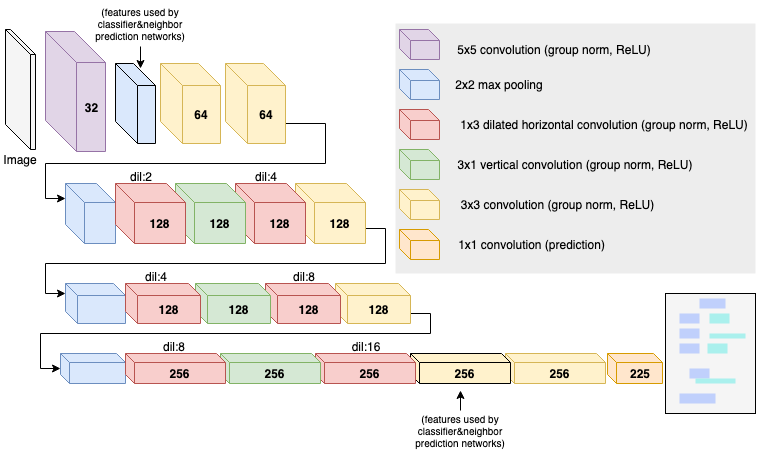}
%includegraphics[width=0.49\textwidth,trim={0 0 0 40},clip]{detectionnet}
\caption{Detector network architecture. Numbers on the boxes are the number of output channels. Dilation amount is indicated above red boxes.
%\vspace{-0.1in}
}
\label{fig:detectornet}
\end{figure}

Using a standard convolutional network (VGG-like) with $3 \stimes 3$ convolutions yields poor results on long text lines (see Fig.~\ref{fig:ex_detect}a). 
For a correct detection, information from the ends of the line must propagate to its center (where the prediction is made). %; otherwise, the width estimation will be inaccurate. 
Thus the lengths of bounding boxes that can be accurately predicted are limited by the horizontal receptive field of the network. 
Although many object detection methods~\cite{yolov3,Girshick_2015_ICCV} use multi-scale approaches, a text line is not at a different scale than the other instances on the page just because it is longer. 
We instead increase the receptive field horizontally by introducing horizontal dilatation \cite{oord2016wavenet}. 
%It has also been observed that a $1 \stimes 3$ convolution followed by a $3 \stimes 1$ convolution has much of the same effect as a $3 \stimes 3$ convolution, while taking less parameters \cite{where is this from?}. 
It can be observed that a $1 \stimes 3$ convolution followed by a $3 \stimes 1$ convolution has much of the same effect as a $3 \stimes 3$ convolution while taking fewer parameters (spatially separable convolution).
As we need only the horizontal increase in our receptive field, we use dilated $1 \stimes 3$ convolutions and non-dilated $3 \stimes 1$ convolutions. 
We apply group normalization~\cite{groupnorm} and ReLU activations between each convolution (we don't use spatially separable convolution, strictly speaking). 
Fig.~\ref{fig:ex_detect} shows a qualitative comparison of results, and Fig.~\ref{fig:detectornet} elaborates our architecture.
%Our architecture can be seen in Fig.~\ref{fig:detectornet}, and
%a qualitative comparison of the results can be seen in Fig.~\ref{fig:ex_detect}.

% \begin{figure}
%     \centering
%     %\begin{subfigure}[]{}
%     \subfloat[Using normal $3 \stimes 3$ convolutions]{
%         \includegraphics[width=0.49\textwidth]{detect_normal}
%     }
    
%     \subfloat[Using dilated $1 \stimes 3$ convolutions]{
%         \includegraphics[width=0.49\textwidth]{detect_staggered}
%     }
%         %\caption{A tiger}
%         %\label{fig:tiger}
%     %\end{subfigure}
%     \caption{Detection with $3 \stimes 3$ convolution (a) and using dilated $1 \stimes 3$ convolutions (b). Blue boxes are pre-printed text detections, cyan boxes are input text detections. Notice in (a) that long lines are either broken into multiple detections (red arrows) or missed.
%     %\vspace{-0.14in}
%     }
%     \label{fig:ex_detect}
% \end{figure}

We used 25 anchor boxes found using $k$-nearest neighbors across the ground truth bounding boxes as in~\cite{yolov2}. 
Our seed points were chosen to span the training/validation distribution via manual inspection.
The training loss is the same as~\cite{yolov3}, except we increase a loss weight by 20 to further encourage incorrect detections to have 0 confidence.
We found this increases model precision.
To prune spurious detections before pairing, we threshold at 0.5 confidence and apply non-maximal suppression.
%We use a threshold (0.5 confidence) and non-maximal suppression of the detections before pairing.

For our optimization we also have the detector network additionally provide a preliminary prediction of the number of neighbors (relationship pairs) for each detected text. 
This is done by having the final $1\stimes 1$ convolution predict an additional value trained with mean-squared-error loss.

%%%%%%%%%%%%%%%%%%%%%%%%%%%%%%%%%%%%%%%%%%%%%%%%%%%%%%%%%%%%%%%%%%%%%%%%%%%%
\section{Pairing}\label{sec:pairing}

Once we have identified pre-printed text and input text lines, we pair them to find label-value relationships.  
First, we identify a high-recall list of potential relationships using a simple heuristic.  
Then, we extract features for each candidate relationship and predict how likely those elements are to be related.  
Finally, because there can be local ambiguity, we use these pairwise scores in a global optimization to derive the final set of relationships.

%\subsection{Candidate Selection}
\subsection{Identifying Candidate Relationships}
\begin{figure}[t]
\centering
\includegraphics[width=0.4\textwidth]{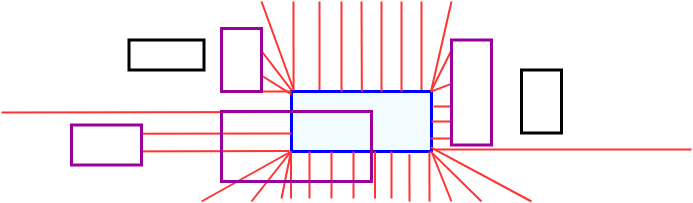}
\caption{Heuristic relationship overprediction is done using line-of-sight. The red lines are the rays determining the line-of-sight of the blue box. Purple boxes are pairing candidates for the blue box.
%\vspace{-0.125in}
}
\label{fig:rays}
\end{figure}

We first identify candidate relationships from the detection results to reduce computation compared to exploring all possible detection pairs.
All pairs of bounding boxes whose edges are within line-of-sight of each other, and are not too far away from each other, are considered candidates. 
The line-of-sight is determined by tracing rays from points along the edges of bounding boxes which terminate after entering a bounding box (see Fig.~\ref{fig:rays}).
To address memory limitations during training, the combined number of candidates and relationships is limited to a pre-determined maximum 
(set to 370 in our implementation).
If the number exceeds the threshold, the maximum length of the rays is shortened and the process is repeated. 
This heuristic has 96.6\% recall for the test set relationships.

%In scene parsing. such a heuristic would be unwise unless the 3D scene were being used. But for forms, where overlap is uncommon, this heuristic is effective.

\subsection{Classifying Candidate Relationships}
\label{sect:classifying}

Many forms place labels to the left of their corresponding value, though sometimes the label may be above or below the value. 
A prior work~\cite{hirayama2011development} attempted to leverage regularity of form layouts by hand-crafting heuristic rules to score potential relationships, but hand-tuned scoring functions 
%based only on distance or simple spatial relationships 
can fail in the template-free case when form layouts do not always match the assumptions made by the heuristic. %, hindering generalization. 
A more generalizable approach is to learn implicit rules by training on a variety of different form layouts.
We use the following features when pairing two element bounding boxes: 
\begin{itemize}
    \item Difference of center $x$ and $y$ positions
    % \item Difference of center $y$ position
    \item Distance from each corner to its counterpart (top-left to top-left, bottom-left to bottom-left, etc.)
    \item Normalized height and width of each bounding box (divided by 50 and 400, respectively)
    % \item Normalized width of each bounding box (divided by constant)
    \item Detector predicted probabilities of belonging to the pre-printed text / input text classes for both bounding boxes
    % \item Predicted probability of belonging to the input text class by detector for each bounding box
    \item Predicted number of neighbors for each bounding box
\end{itemize}

\begin{figure}[t]
\centering
\includegraphics[width=0.48\textwidth]{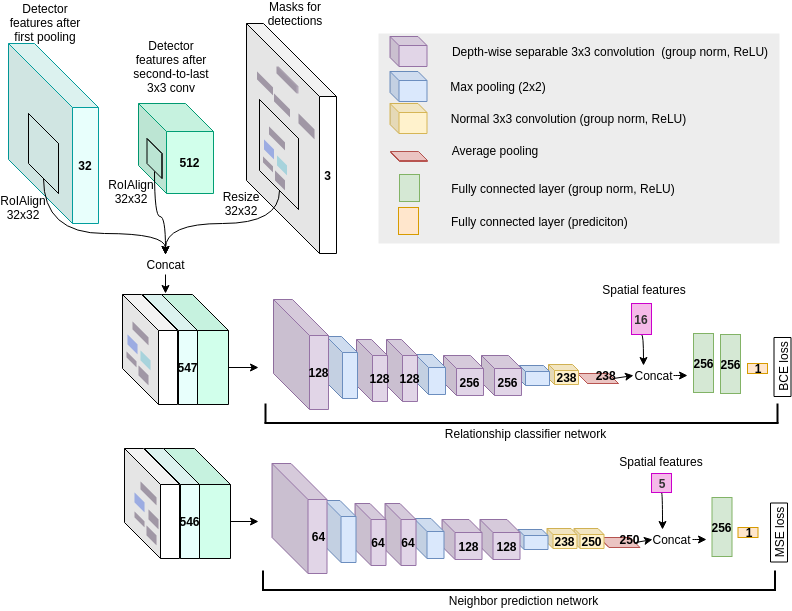}
\caption{Relationship classifier and neighbor prediction networks. 
Numbers on the boxes indicate the number of output channels.
Both networks receive as input the concatenation of the detector's first pooling layer and second-to-last convolution layer (extracted with RoIAlign \cite{roialign}), as well as resized detection masks.
%Resized masks of the location of masks is also appended. 
The relationship classifier predicts the probability that the two input detections have a relationship.
The neighbor prediction network refines the prediction of the number of neighbors for the input detection.
We use depth-wise separable convolution to reduce the number of parameters \cite{howard2017mobilenets}. 
%\vspace{-0.125in}
}
\label{fig:net}
\end{figure}

It is clear that, in addition to spatial features, humans also use multiple visual cues in determining relationships: lines, borders, nearby text and handwriting, etc. 
To allow for the learning of these cues, we also use a convolutional network to analyze the area surrounding each potential pairing.

For each candidate relationship we find the rectangular area bounding the two bounding box detections and pad it by 150 pixels on each side to provide local context. 
We append detector network features from both the second-to-last convolution layer and the first pooling layer.
These features are cropped with RoIAlign \cite{roialign} to the size $32 \stimes 32$. 
We append three additional binary masks to these features (resized to $32 \stimes 32$), one mask for each bounding box in the candidate relationship and a mask of all detected bounding boxes (Fig.~\ref{fig:net}, top left).
The input order of the candidate text bounding box masks are randomized during training. 
For evaluation we average the result of both orderings.
%While the features are taken from the detection network, and thus contain information regarding the objects, the masks provide the network with two important things: 
%1)~the output of the non-maximal suppression, and 
%2)~explicit features with which it can reason about all detections in the neighborhood.

We extract features from this input tensor with a small convolutional network and apply global pooling to the resulting features.
The resulting flattened features are appended to the previously described spatial features, and a fully connected network classifies the candidate relationship as valid or not.
This network is trained with a binary cross-entropy loss. 
Fig.~\ref{fig:net} shows the pairing network architecture.
%Appending the spatial features allows the network to use a shape and location prior, similar to \cite{zhang2017relationship}, but more learned since we do not directly classify only on the spatial information.

%%%%%%%%%%%%%%%%%%%%%%%%%%%%%%%%%%%%%%%%%%%%%%%%%%%%%%%%%%%%%%%%%%%%%%%%%%%%
\section{Neighbor Prediction Network}

For the subsequent global optimization (Section \ref{sec:opt}), we predict the number of neighbors (relationships) each detected printed/input text element has. 
While the detection network makes initial predictions, better predictions can be made after removing spurious predictions and by focusing on each individual text detection.

We apply another small convolutional network (Fig.~\ref{fig:net}) to the region around each detection, in a similar manner as described in Section \ref{sect:classifying}.
However, this network has only two input masks: one for the detection of interest and one for all detections. 
The features appended before the fully-connected layers are also slightly different: normalized height and width, initially predicted number of neighbors, and class prediction.

%%%%%%%%%%%%%%%%%%%%%%%%%%%%%%%%%%%%%%%%%%%%%%%%%%%%%%%%%%%%%%%%%%%%%%%%%%%%

% *** placed here to force placement after next page break ***
\begin{table*}[t]
\caption{NAF Dataset characteristics}
\label{table:dataset}
\vspace{-0.05in}
\centering
\begin{tabular}{llcccccc}
 Version & Split                      & Images  & \# Form Types & Pre-printed Text & Input Text & Label-Value relationships   \\
\hline        
\multirow{3}{*}{Simple}& Train                 & 143 & 51   & 4547 & 2589 & 2496 \\
& Validation      & 11 & 6     &  368 & 162 & 159        \\
&Test         & 11 & 8     &  250 & 189 & 161        \\
\hline
\multirow{3}{*}{Full} & Train                & 682 & 209    & 40347 & 12482 & -\\%*14356 \\
&Validation      & 59 & 31     &  3381 & 1266 &  -\\%*1531       \\
&Test         & 63 & 34     &  2892 & 1229 &  -\\%*1423       \\
\end{tabular}
\vspace{-0.1in}
\end{table*}

\section{Global Optimization}\label{sec:opt}

The problem of determining a single relationship can depend on other relationship decisions for a form. 
Imagine the scenario where a pre-printed text line has an equal probability to be in a relationship with two different handwriting instances, but one of the handwriting instances has another possible pairing and the other does not. 
Assuming we know each of these handwriting instances should be paired with only one pre-printed text instance, we can easily recognize the appropriate pairings for them.
To take all relationship decisions into account at once we employ global optimization as a post-processing step.

Ideally, we want to encourage the number of predicted relationships per bounding box to be similar to the predicted number of neighbors for each detected element, while respecting the probability or score of the relationships.

%Let~$X$ be a vector of binary labels $x_{r}$ indicating whether a relationship $r$ is included, $p_{r}$ be the network's probability that the relationship exists, and $n_i$ be the estimated number of neighbors for the $i^{th}$ detected bounding box. 
%Let $R_i$ be the set of relationships $x_{r}$ that include the $i^{th}$ detected bounding box.
%$c$ is a tune-able parameter that determines how much confidence we place in the accuracy of $n_i$.
Let~$R$ be the set of candidate relationships and~$\mathbf{x}$ be a vector of binary labels, such that $x_r=1$ indicates that relationship $r \in R$ is accepted.
Let $p_r$ be the pairing network's predicted probability for $r$,
%$x_{r} {\in} X$ indicating whether a relationship $r {\in} R$ is included, and let $p_{r}$ be the network's probability that the relationship exists.
$n_b \in \mathbb{R}$ be the estimated number of neighbors for the detected bounding box $b \in B$,
and $R_b \subseteq R$ be the subset of all relationships that $b$ is part of.
The tune-able parameter $c$ determines how much confidence we place in the accuracy of $n_b$, and $T$ is a (soft) threshold.
We formulate our optimization as
\begin{equation}
\mathbf{x^*}\! = \! \underset{\mathbf{x}}{\text{argmax}}\!
\left[
%\sum\limits_{r=1}^R (p_{r}-0.5)x_{r} - c\sum\limits_{i}^B (n_i - \!\!\sum\limits_{x_{r} \in R_i} x_{r})^2
\sum\limits_{r \in R} (p_{r}\!-T)x_{r}\! - c\sum\limits_{b \in B} \big(n_b - \!\!\sum\limits_{r \in R_b} x_{r}\big)^2
\right]
\label{eq:optimization}
\end{equation}
%subject to $x_{r} \in \{0,1\}$.
%&&& x_{i,j} = 0 \text{ where pair not found by heuristic}

The first term of Eq.~\ref{eq:optimization} seeks to reject relationships with probabilities less than $T$. 
With $c=0$, Eq.~\ref{eq:optimization} reduces to thresholding with $p_r \geq T$. 
The second term regularizes each $b$ to have $n_b$ neighbors.
To handle uncertainty, $n_b$ can be a non-integer.
For example, if $b$ could have 0 or 1 neighbors, having $n_b=0.5$ equally penalizes both cases.
We found $c=0.25$ and $T=0.7$ worked well on the validation set for most experiments and used this in our evaluation.
We use the branch-and-bound variant of ECOS~\cite{Domahidi2013ecos} to solve Eq.~\ref{eq:optimization}.

%%%%%%%%%%%%%%%%%%%%%%%%%%%%%%%%%%%%%%%%%%%%%%%%%%%%%%%%%%%%%%%%%%%%%%%%%%%%

\section{NAF Dataset}\label{sec:dataset}

We introduce and release a new dataset of annotated historical form images, the National Archives Forms (NAF) dataset, with the following properties:
\begin{itemize}
    \item Varied form layouts, with train, validation, and test sets having disjoint form layouts.
    \item Historical, noisy.
    \item Filled in by hand and/or typewriter.
\end{itemize}

\begin{figure*}
    \centering
    %\begin{subfigure}[]{}
    % \subfloat[]{
    %     \includegraphics[width=0.36\textwidth]{007675712_00119}%AP:0.542
    % }
        
    % \subfloat[]{
    %     \includegraphics[width=0.38\textwidth]{100093596_00222} %AP:0.302
    % }
    
    % \subfloat[]{
    %     \includegraphics[width=0.18\textwidth]{101023522_00018} %AP:0.491
    % }
    \includegraphics[width=1\textwidth]{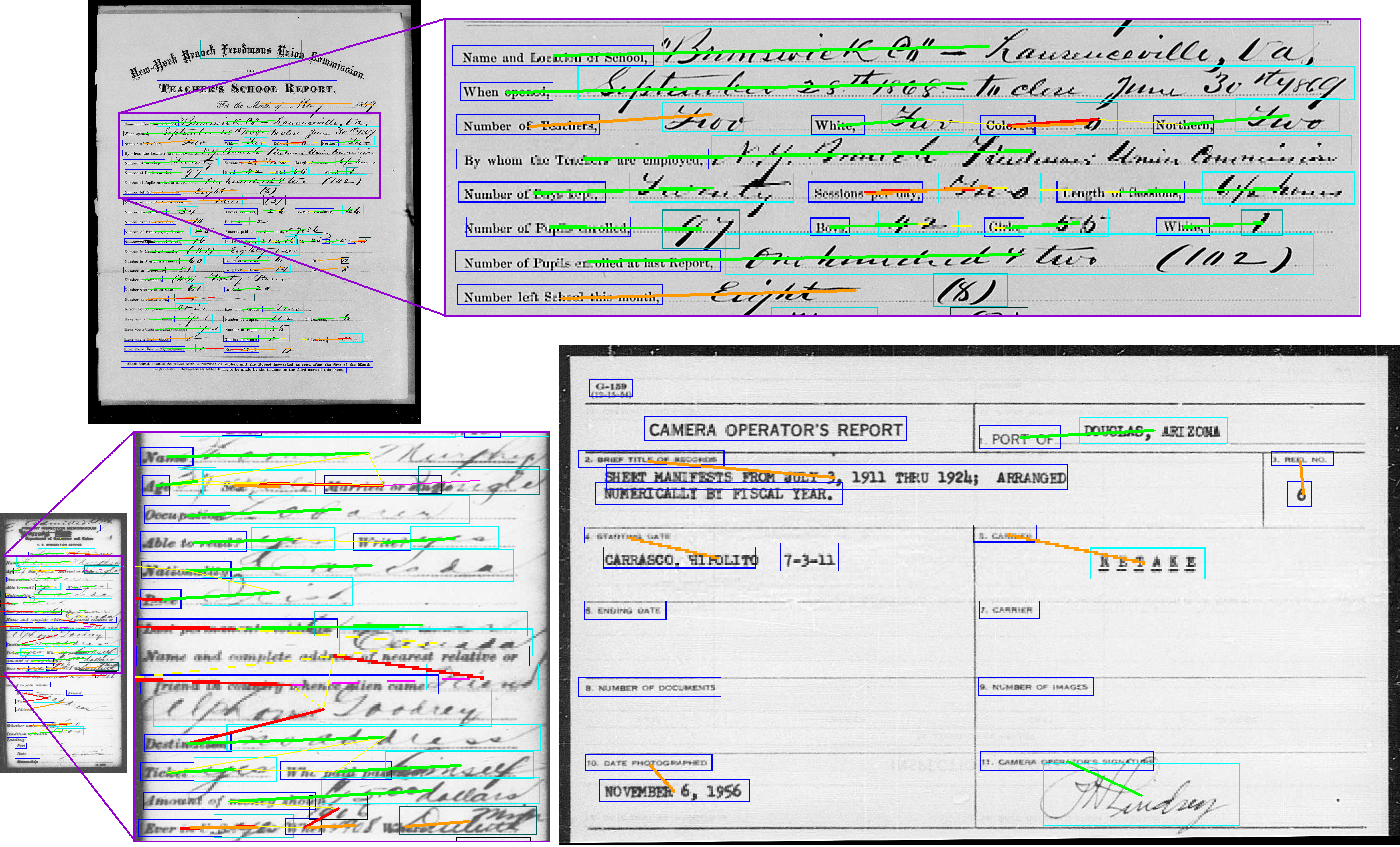}
       
    \caption{Test set examples. Blue and cyan boxes are pre-printed text and input text detections respectively. Green lines are correct relationship predictions, red lines are false positive errors, orange lines are false negative errors, thin yellow lines are relationships correctly pruned by the optimization, thin pink lines are relationships incorrectly pruned by the optimization. The relationship AP of the images: top-left 0.625, bottom-left 0.503, bottom-right 0.371
    %\vspace{-0.05in}
    }
    \label{fig:pairing_results}
\end{figure*}

The NAF dataset is comprised of historical form images from the United States National Archives.
%\footnote{\url{https://www.archives.gov/}}. 
The images are noisy due to degradation and the machinery used to print them.
Figures \ref{fig:top_fig}, \ref{fig:ex_detect}, and \ref{fig:pairing_results} contain examples from the dataset.

We have restricted our pairing dataset to images not containing tables or prose/fill-in-the-blank information in order to focus on the label-value problem (other approaches will be more effective for these types of forms). 
However, we use the full dataset for pre-training the detection network.

We divided the images into training, validation, and test sets, where each set has a distinct set of form layouts, though there are multiple instances of each form layout within each set.
This mimics the template-free scenario, i.e., we test on form layouts our system has never seen before. 
Details of this dataset can be seen in Table~ \ref{table:dataset}. 
%The form types in the validation and test sets are each distinct from the other types in the set, while the training set contains some types that are only slight variations of eachother.

%The validation and test set have each have 6 form types.

%Information on a form is generally presented in one of four formats
%\begin{itemize}
%    \item Label-value: Some text indicates what information is to be recorded and then a nearby field exists for the information to be recorded in. See X in Fig. X.
%    \item Tables: Generally columns are headed to identify what information is to be recorded and rows represent different instances which are being recorded. Lists can be viewed as special cases of tables with a single column. %See X in Fig. X?
 %   \item Prose/fill-in-the-blank: Sentences of prose are written with blanks that are to be filled in. Ex: ``I \underline{  } do hereby grant the property at \underline{  } to \underline{  }...'' %See X in Fig. X?
%    \item Comments: This is information that the form was not originally intended to capture, but added by the person filling in the form. See X in Fig. X.
%\end{itemize}

Elements of the images are annotated with quadrilaterals, which we convert to axis-aligned rectangles. 
%The elements that are annotated are pre-printed text, input text/handwriting, and figures/pictures, with specific labels/classes (label, field, sub-label, prose/fill-in-the-blank, etc.). 
%Tables are specially labeled. The elements are also paired if a relationship exists. 
For this work, we use only the pre-printed text and input text annotations, though the dataset does contain richer annotations. %, and only use relationships between opposite classes. 
%The annotated elements are paired with each other if a relationship exists between them. %(the relationship being inferred from the classes of the elements). 
We use only the relationships between pre-printed text and input text elements as these typically are label-value relationships.
%The text and handwriting are not yet transcribed.

%There is some information captured by the person filling in a form either circling or crossing out pre-printed text. We currently ignore this type of information

\section{Experiments and Results}
\label{sec:results}

\subsection{Training}
We train all our models with the Adam optimizer~  \cite{kingma2014adam} and used the validation set to determine %the number of iterations to run our model and other 
hyper-parameters.

For both detection and pairing, we uniformly randomly resize training images to 0.4--0.65 of their original size. 
A~training instance is a random $652 \stimes 1608$ crop of the resized image. 
This size captures several complete relationships while using less memory than a full image. 
If a text instance is cropped horizontally, we clip its bounding box to fit in the window. 
If a text instance is cropped vertically so that less than half the bounding box is inside the window, we remove the instance from the ground truth. %otherwise, we leave it unmodified, expecting the network to predict the handwriting/font height.
For data augmentation, we also randomly perturb contrast as in~\cite{wigington2018start}.

\textbf{For detection:}
We use the full dataset for training the detector network. 
We apply additional data augmentation when training the detection network by randomly rotating images slightly and flipping them horizontally. We use a learning rate of 0.01 and a batch size of 5.
%The models we present results for are trained for 200,000 iterations.
We pre-train the detector network to 150,000 iterations.

\textbf{For pairing:}
%We use a subset of the dataset that does not include forms with tables and fill-in-the-blank prose.
We use a subset of the dataset as described in Section \ref{sec:dataset}.
The detector network is frozen for the first 2,000 iterations and afterwards %is trained with the classifier and neighbor prediction networks.
its weights are fine-tuned through all tasks' losses.
At each training iteration for the pairing network we present either  predicted or ground-truth bounding boxes. 
The probability of using ground truth bounding boxes is initially 100\% and then lowered until it reaches 50\% at 20,000 iterations. 
We use an IoU threshold of 0.4 in aligning predicted and ground truth bounding boxes to determine which predicted relationships are true. 
We threshold detections  at 0.5 IoU. 
If a predicted bounding box does not overlap with any ground truth, all possible relationships with it are false. 
If a prediction overlaps with ground truth by less than the IoU threshold, we do not calculate the loss for its relationships that would be true.
We use a learning schedule similar to \cite{vaswani2017attention} with a warm-up of 1,000 iterations, a maximum learning rate of 
%$1.5 \cdot 10^-3$ 
0.0015 
and a mean learning rate of
0.00062
%$6.2 \dot 10^-4$
. The batch size is 1. We terminate training at 125,000 iterations.
%To balance tasks in training, w
We weight the multiple loss terms as follows, detection: 1.0, pairing: 0.5, and number of neighbor regression: 0.25. 
%The detection loss has the highest weight so what it has learned is not forgotten in training the pairing.

\subsection{Evaluation}

Qualitative detection and pairing results for images from the test set can be seen in Fig.~\ref{fig:pairing_results}. It can be observed in the top image crop and bottom-left image crop that the optimization removes several false-positive relationships which are not consistent with neighboring predictions (thin yellow lines). In the bottom-right image the detector struggles to correctly predict the class of input text that is printed, and thus misses several relationships. Other observations we made of the results are that the model struggles with predicting long distance relationships, and continues to have errors where multiple relationships are plausible, even after optimization. For many errors it is unclear what the cause is, e.g.~the model predicts the correct number of neighbors for two detections that should be paired, but predicts a low probability of pairing them in the absence of obvious distractors.

For quantitative evaluation we measure mean average precision (mAP), recall, precision, and F-measure (F-m) for both detection and relationship predictions. 
For a text detection to be correct it must have at least 0.5 IOU with a GT bounding box and match the GT class. For a predicted relationship to be correct it must be between two correct detections whose matched GTs have a relationship.

\begin{table*} % placed here just to put the tables together
\caption{Detection results}
\centering
\label{table:detection_results}
\vspace{-0.1in}
\begin{tabular}{l|c|cccccc|cc}
                                & & \multicolumn{6}{c|}{Pairing dataset} & \multicolumn{2}{c}{Full dataset} \\
                                &           &     &  avg. &  \multicolumn{2}{c}{pre-printed text} & \multicolumn{2}{c|}{input text}& & avg.   \\
Method                          & \# params  & mAP & F-m & prec.  &  recall & prec. & recall & mAP & F-m   \\
\hline
Standard ConvNet (VGG-like)               & 3M & 0.364 & 0.719 &  0.811 & 0.780 & 0.689 & 0.603   &   0.324 & 0.612        \\
%Staggered convs (w/ $1\stimes 1$ convs)      & 6.8M & 0.415 & 0.435 & .352            \\
Dilated staggered convs  (Fig.~\ref{fig:detectornet})  pre-trained     & 2.4M & 0.423  & \textbf{0.836} &  \textbf{0.861} &\textbf{0.908} &  \textbf{0.816}   &0.763  & \textbf{0.421} & \textbf{0.808} \\
Dilated staggered convs (Fig.~\ref{fig:detectornet}) fine-tuned      & 2.4M & \textbf{0.428}  & 0.795 &  0.791 &0.906 & 0.726  &\textbf{0.768}  &  - & -
\end{tabular}
%\vspace{-0.05in}
\end{table*}

% \begin{table*}
% \caption{Pairing results}
% \label{table:pairing_results}
% \vspace{-0.1in}
% \centering
% \begin{tabular}{l|cccc|cccc}
%                          &  \multicolumn{4}{c|}{Without optimization}  & \multicolumn{4}{c}{After global optimization}  \\ 
% Method                         & mAP & F-m & prec. & recall & mAP & F-m & prec. & recall  \\ 
% \hline
% Distance based                          & 0.242 & 0.296 & 0.191 & 0.651   & 0.258 & 0.343 & 0.256 & 0.519     \\
% Scoring functions from \cite{hirayama2011development} & 0.213 & 0.113 & 0.233 & 0.091   & 0.214 & 0.157 & 0.251 & 0.114          \\
% Classifier w/o visual features  & 0.293 & 0.257 & 0.160 & 0.651  & 0.368 & 0.370 & 0.267 & 0.603           \\
% ConvNet with visual features      & \textbf{0.549}  & \textbf{0.493} & \textbf{0.394} & \textbf{0.661}   & \textbf{0.554} & \textbf{0.552} & \textbf{0.476} & \textbf{0.658} 
% \end{tabular}
% %\vspace{-0.05in}
% \end{table*}

%NEW RESULTS usign 0.7 threshold
\begin{table*}
\caption{Pairing results}
\label{table:pairing_results}
\vspace{-0.1in}
\centering
\begin{tabular}{l|cccc|cccc}
                         &  \multicolumn{4}{c|}{Without optimization}  & \multicolumn{4}{c}{After global optimization}  \\ 
Method                         & mAP & F-m & prec. & recall & mAP & F-m & prec. & recall  \\ 
\hline
Distance based                          & 0.235 & 0.217 & 0.134 & 0.666  & 0.251 & 0.306 & 0.254 & 0.428     \\
Scoring functions from \cite{hirayama2011development} & 0.135 & 0.063 & 0.162 & 0.080   & 0.136 & 0.077 & 0.151
& 0.086          \\
Classifier w/o visual features  & 0.248 & 0.240 & 0.157 & 0.680  & 0.275 & 0.352 & 0.277 & 0.516           \\
ConvNet with visual features      & \textbf{0.585}  & \textbf{0.589} & \textbf{0.559} & \textbf{0.655}   & \textbf{0.584} & \textbf{0.607} & \textbf{0.654} & \textbf{0.599} 
\end{tabular}
%\vspace{-0.05in}
\end{table*}

% \begin{table}[t]
% \caption{Experiment results given perfect detection }
% \label{table:ablation_results}
% \vspace{-0.1in}
% \centering
% \begin{tabular}{l|cc|cc}
%                                 & \multicolumn{2}{c|}{optimized} &  \multicolumn{2}{c}{using GT} \\ 
%                                 & \multicolumn{2}{c|}{w/ GT NN}  &  \multicolumn{2}{c}{detections} \\ 
% Method                          & mAP  &  F-m & mAP  &  F-m\\ 
% \hline
% Distance based                           & 0.296 &  0.471   & 0.457 & 0.394     \\
% Scoring functions from \cite{hirayama2011development} & 0.215 & 0.155   & 0.364 & 0.180         \\
% Classifier w/o visual features   & 0.395 &  0.442  & 0.608 & 0.447           \\
% ConvNet with visual features       & \textbf{0.642} & \textbf{0.697}   & \textbf{0.876} & \textbf{0.788}
% \end{tabular}
% %\vspace{-0.125in}
% \end{table}

%NEW DATA using 0.7 threshold
\begin{table}[t]
\caption{Upper bound experiments using perfect information}
\label{table:ablation_results}
\vspace{-0.1in}
\centering
\begin{tabular}{l|cc|cc}
                                & \multicolumn{2}{c|}{optimized} &  \multicolumn{2}{c}{using GT} \\ 
                                & \multicolumn{2}{c|}{w/ GT NN}  &  \multicolumn{2}{c}{detections} \\ 
Method                          & mAP  &  F-m & mAP  &  F-m\\ 
\hline
Distance based                           & 0.413 &  0.504  & 0.424 & 0.314     \\
Scoring functions from \cite{hirayama2011development} & 0.136 & 0.073   & 0.238 & 0.085         \\
Classifier w/o visual features   & 0.509 &  0.597  & 0.428 & 0.377           \\
ConvNet with visual features       & \textbf{0.640} & \textbf{0.721}   & \textbf{0.912} & \textbf{0.855}
\end{tabular}
%\vspace{-0.125in}
\end{table}

%F-measure requires a decision to be made regarding accepting or rejecting potential relationships, if we are optimizing, the optimization makes this decision, otherwise we use a threshold of X. 
Average precision (AP) requires continuous scores, so when we optimize we subtract 1 from the probability of each rejected relationship to maintain order before calculating AP. For F-m, recall, and precision we threshold the detector at 0.5, and we threshold pairing at 0.5 ($T{=}0.5$ for optimization).

We first compare our detection network architecture to a standard convolutional (VGG-like) network that does not use dilation. 
%Additionally, we compare against an architecture that only uses $3 \time 1$ and $1 \time 3$ kernels with additional $1 \time 1$ convolutions after each pair to validate the mixed architecture we used. 
Table \ref{table:detection_results} shows the number of parameters in each model and their respective performance on the full test set (average of five different training runs). 
While the dilated architecture we propose has fewer parameters, it significantly outperforms the standard convolutional network. 

To demonstrate that the proposed learning-based pairing method outperforms simple heuristics, we implemented two baseline methods based on heuristic rules. % from~\cite{hirayama2011development}.
The first is a simple one based on inverse distance (i.e.~closer elements of opposite classes are more likely to be paired):
%For this simple distance rule the score for the relationship between pre-printed text detection~$i$ and input text detection~$j$ is
%\begin{equation*}
\begin{equation}
%s_{i,j} = 1- (dist( (x_i,y_i), (x_j,y_j) )-d_{min})/(d_{max}-d_{min})
%s_{i,j} = 1- \frac{dist( (x_i,y_i), (x_j,y_j) )-d_{min}}{d_{max}-d_{min}}
s_{i,j} = 1- \frac{\left\| (x_i,y_i) -  (x_j,y_j) \right\| -d_{min}}{d_{max}-d_{min}}
    %\begin{cases} 
    %  1- (dist( (x_i,y_i), (x_j,y_j) )-d_{min})/(d_{max}-d_{min}) & \text{if $i$ and $j$ is a potential %relationship}\\
    %  0 & \text{otherwise}
    %  
    %\end{cases}
\end{equation}
%\end{equation*}
where $(x_i,y_i)$ and $(x_j,y_j)$ are the centers of the bounding boxes for two detected text elements of different class, and $d_{max}$ and $d_{min}$ are the maximum and minimum distances for all potential relationships.
The second baseline uses a scoring function adapted from~\cite{hirayama2011development}. 
They use cell boundaries (of field areas) in some of their scoring terms, which we cannot use, so we use only the scoring terms based on height, distance, and whether the value is to the right of the label.
\begin{comment}
The scoring we adapt from~\cite{hirayama2011development} is
\begin{equation}
s_{i,j} = 1 - g4(i,j) - g5(i,j) - g6(i,j) 
\end{equation}
\begin{equation}\label{g4}
g4(i,j) = \frac{\max(h_i,h_j) }{ \min(h_i,h_j) }
\end{equation}
\begin{equation}
\delta_{i,j} = \left\| (x_i+w_i/2,y_i) -  (x_j-w_j/2,y_j) \right\| 
\end{equation}
\begin{equation}\label{g5}
g5(i,j) =
    \begin{cases} 
      \delta_{i,j} / w_i & \delta_{i,j} > 2 w_i \\
      \min(1,\delta_{i,j} / w_i) & \text{otherwise}
   \end{cases}
\end{equation}
\begin{equation}\label{g6}
g6(i,j) = (x_i-x_j)/w_j
\end{equation}
where $x_i$ and $y_i$ are the center of detection $i$, and $h_i$ and $w_i$ are its height and width. 
\end{comment}
Intuitively, the scoring penalizes pairing text elements of different heights, % (Eq.~\ref{g4}),
distantly separated text elements, % (Eq.~\ref{g5}), 
and input text to the right of the pre-printed text. % (Eq.~\ref{g6}).

To evaluate the additive effect of using visual cues in addition to  spatial features, we implemented a baseline network that takes as input only the spatial features listed in Section~\ref{sect:classifying} and not the contextual visual features our full method does.
For this classifier we use three fully connected layers with batch normalization, dropout, and ReLU activations for the first 2 layers. We use a hidden size of 256. We trained the network with a binary cross-entropy loss, a learning rate of 0.001, and a batch size of 512 for 6,000 iterations.

The performance of our proposed method %(using both spatial and visual features) 
compared to these baseline methods is shown in Table~\ref{table:pairing_results} (average of 5 training runs). 
%As would be expected, performance increases with the complexity of the method, and the proposed learning-based approach using both spatial and visual features significantly outperforms both heuristic methods and the classifier without visual features. 
Surprisingly the distance-based method's performance is similar to the non-visual  classifier. 
%This indicates that, by themselves, simple non-visual features are not discriminative. 
Including visual features significantly outperforms any of the baselines.
The global optimization improves F-measure as it sacrifices some recall for greater precision. 
The gains are more evident with the distance-based heuristic and the non-visual classifier. 
Our model sees a context window and so can already reason about neighbors without the optimization.

\subsection{Additional Experiments}

%Table~\ref{table:pairing_results} also shows the performance of each method before and after post-processing optimization. 
%For the distance-based heuristic and the non-visual classifier, the post-processing significantly boosts performance. 
%It does not, however, significantly increase the performance of the full (visual) network, likely because the visual context window already allows the network to reason about its neighbors.
As seen in Table~\ref{table:ablation_results} (average of five training runs), substituting the ground-truth number of neighbors during the optimization instead of the predicted number (and using $c=25$) greatly increases the effectiveness of the optimization.

Because this work focuses on pairing form elements,
%and we admit that other methods of text detection \cite{gruning2018,wigington2018start} would likely be superior to ours. 
we also evaluate the upper-bound performance of our proposed and baseline pairing methods by using ground truth text detections instead of predicted ones (Table \ref{table:ablation_results}, average of five training runs).
This allows us to examine the performance of the pairing network independent of the means of detection.
The number of neighbors is part of the detection ground truth; to minimize this information's impact we introduce a $\pm 1$ uniform noise to the number of neighbors.
As expected, all of the methods improve when given perfect detections as input.%, but notably the gap in performance between the proposed method and the baseline methods widens significantly as the accuracy of the detections increases.

We also measured the contribution made by the neighbor prediction network, which refines the predicted number of neighbors after the initial detection network.
%The optimization performance is dependent on the accuracy of the prediction of the number of neighbors. 
The neighbor prediction network predicts the number of neighbors with 72\% accuracy, while the detector alone predicts the number of neighbors with 50\% accuracy,
suggesting that the use of this auxiliary network is helpful.
%This validates our use of the auxiliary network.

%%%%%%%%%%%%%%%%%%%%%%%%%%%%%%%%%%%%%%%%%%%%%%%%%%%%%%%%%%%%%%%%%%%%%%%%%%%%

% An example of a floating figure using the graphicx package.
% Note that \label must occur AFTER (or within) \caption.
% For figures, \caption should occur after the \includegraphics.
% Note that IEEEtran v1.7 and later has special internal code that
% is designed to preserve the operation of \label within \caption
% even when the captionsoff option is in effect. However, because
% of issues like this, it may be the safest practice to put all your
% \label just after \caption rather than within \caption{}.
%
% Reminder: the "draftcls" or "draftclsnofoot", not "draft", class
% option should be used if it is desired that the figures are to be
% displayed while in draft mode.
%
%\begin{figure}[!t]
%\centering
%\includegraphics[width=2.5in]{myfigure}
% where an .eps filename suffix will be assumed under latex, 
% and a .pdf suffix will be assumed for pdflatex; or what has been declared
% via \DeclareGraphicsExtensions.
%\caption{Simulation Results}
%\label{fig_sim}
%\end{figure}

% Note that IEEE typically puts floats only at the top, even when this
% results in a large percentage of a column being occupied by floats.

%\textcolor{red}{BSM: reworked to here}

%%%%%%%%%%%%%%%%%%%%%%%%%%%%%%%%%%%%%%%%%%%%%%%%%%%%%%%%%%%%%%%%%%%%%%%%%%%%
\section{Conclusion}
We have introduced a trainable, language-agnostic method to detect and pair pre-printed text and input text in form images that is robust to noise found in historical documents. 
%Our method uses a specialized FCN architecture for detecting long text lines and a convolutional network to classify candidate relationships. 
%It also uses a post-processing optimization to encourage global consistency across a form.
We have also introduced the NAF dataset, which contains images of historical forms with a variety of layouts, and evaluated our method against alternative baselines using this dataset. 
%The proposed method outperforms heuristic baselines.
There is not an existing benchmark for this problem.

The results presented here show that dilated $1\stimes 3$ convolutions make a FCN more effective at detecting long text lines. 
These results also indicate that having a learned method that uses visual features is important when pairing text lines.
We have also found that optimizing results across a page leads to increased precision.

%The proposed method outperforms heuristic baselines.
%Additional experiments show that the use of visual features significantly increases performance compared to using spatial features alone.

% conference papers do not normally have an appendix

% use section* for acknowledgement
\section*{Acknowledgment}

We would like to thank the United States National Archives and FamilySearch for the images used in the NAF dataset. We also acknowledge the impact of Bill Barrett's formative ideas on this work.

% trigger a \newpage just before the given reference
% number - used to balance the columns on the last page
% adjust value as needed - may need to be readjusted if
% the document is modified later
%\IEEEtriggeratref{8}
% The "triggered" command can be changed if desired:
%\IEEEtriggercmd{\enlargethispage{-5in}}

% references section

% can use a bibliography generated by BibTeX as a .bbl file
% BibTeX documentation can be easily obtained at:
% http://www.ctan.org/tex-archive/biblio/bibtex/contrib/doc/
% The IEEEtran BibTeX style support page is at:
% http://www.michaelshell.org/tex/ieeetran/bibtex/
%\bibliographystyle{IEEEtran}
% argument is your BibTeX string definitions and bibliography database(s)
%\bibliography{IEEEabrv,../bib/paper}
%
% <OR> manually copy in the resultant .bbl file
% set second argument of \begin to the number of references
% (used to reserve space for the reference number labels box)
{ \small
\bibliographystyle{IEEEtran}
\bibliography{bib}
}

% that's all folks
\end{document}